Paper number XXX-XXXX

# Mining Personalized Climate Preferences for Assistant Driving

**Feng Hu[1,2], Ning Duan[1], Jingchang Huang[1], Wei Sun[1], Zhigang Zhu[3,2]**
1. IBM Research – China
2. The Graduate Center, The City University of New York
3. The City College, The City University of New York

## Abstract

Both assistant driving and self-driving have attracted a great amount of attention in the last few years. However, the majority of research efforts focus on safe driving; few research has been conducted on in-vehicle climate control, or assistant driving based on travellers' personal habits or preferences. In this paper, we propose a novel approach for climate control, driver behavior recognition and driving recommendation for better fitting drivers' preferences in their daily driving. The algorithm consists three components: (1) A in-vehicle sensing and context feature enriching compnent with a Internet of Things (IoT) platform for collecting related environment, vehicle-running, and traffic parameters that affect drivers' behaviors. (2) A non-intrusive intelligent driver behaviour and vehicle status detection component, which can automatically label vehicle's status (open windows, turn on air condition, etc.), based on results of applying further feature extraction and machine learning algorithms. (3) A personalized driver habits learning and preference recommendation component for more healthy and comfortable experiences. A prototype using a client-server architecture with an iOS app and an air-quality monitoring sensor has been developed for collecting heterogeneous data and testing our algorithms. Real-world experiments on driving data of 11,370 km (320 hours) by different drivers in multiple cities worldwide have been conducted, which demonstrate the effective and accuracy of our approach.

## Introduction

An autonomous car, which is capable of sensing its environment and navigating without human input, has more and more become a reality for general public nowadays. An Advanced Driver Assistance System (ADAS) aims to provide assistance for automating, adapting or enhancing a vehicle system for safety and better driving instead of fully taking over of the control driving. The major focus of the current ADAS systems (or sometimes called self-driving systems), however, is to collect and interpret various sensory information to identify appropriate and safe navigation paths, as well as avoiding obstacles and observing relevant traffic signage. Few research has been conducted on identifying drivers' in-vehicle climate preference patterns, correlating with their surroundings, and providing automatic climate control for bettering user experience beyond safety.

We have developed a vehicle surrounding information collecting platform equipped with heterogeneous sensing data both inside and outside of a vehicle on the road. By identifying drivers' behaviour patterns and the relationships with their surroundings -- taking driving in heavily polluted cities as an example, we have trained models for individual drivers that further provide customized action suggestion, e.g., close/open windows, control air condition,, which could be integrated into next generation of self-driving vehicles.

The contributions of our work can be summarized as the following: (1) An integrated IoT platform is proposed based on parameters of environments, vehicles, and traffics. (2) An intelligent vehicle status recognition algorithm is developed, which can automatically judge vehicle's current window and air-conditioning status by monitoring the changes of the multiple sensor data. (3) A novel driver preference identification approach for providing the best driving recommendations.

The organization of the rest of the paper is as follows. Section 2 discusses related work. Section 3 explains the main idea of the proposed solution. Section 4 provides real data experimental results. Finally, Section 5 gives a conclusion and points out some possible future research directions.

**Related Work**

Researchers have demonstrated that in-vehicle air pollution is improvable by utilizing a commercial air purifier [1], an assembled mobile sensing box [2], or an intelligent integrated air quality management system [3]. However, few further analysis has been carried out to find the correlation between the vehicle air quality and surrounding influencing factors, such as temperature, humidity, etc., based on which individual driver preferences can be distinguished and identified.

Hidden Markov Model (HMM) is a statistical model widely used in systems that are assumed to be Markov processes with unobserved states [4][10-11]. Deep learning is another branch of machine learning techniques that attempts to model high level abstractions in a large dataset by using a deep graph with multiple processing layers composed of multiple linear and non-linear transformations [12]. However, both of these two approaches require a large amount of data to train an accurate and robust model. In our application, since we are customizing each driver's behaviour pattern and preference, it is impractical to request each driver to acquire a large amount of driving data that can satisfy the need to train a large neural network model.

Q-learning is a model-free reinforcement learning technique which can be used to find an optimal action-selection policy for any given Markov decision process [13-14]. Q-learning is preferable in tasks where the goal is to achieve optimal response in any given states. However, in our task, in many cases, drivers' actions are subjective, and there are no optimal actions available (e.g. the preferable in-vehicle temperature or humidity).

Self-driving vehicle has achieved significant development from both theory and technology aspects within both academic and industrial communities. A classical self-driving system's architecture includes hardware, a car operating system, middle ware and an application layer [15, 16]. The current popular vehicle applications are mostly for functionalities, such as lane detection, lane following, overtaking, and parking. Very few attention, however, has been paid to the relationship between the behaviors (e.g. open/close windows) of the driver and their surrounding environment (e.g., temperature, air quality index). A user's traveling experience would be a detriment of comfort if the vehicle leaves its the windows opened all the time when the surrounding air quality is highly polluted, or keeps the windows closed all the time without letting in fresh air. Motivated by these user driving experience concerns, we propose a novel personalized self-driving climate preference mining approach, and provide a highly customized driving climate control system for enhancing user experience.

## Algorithms

Customizing drivers' climates needs to acquire the drivers' actions under different surrounding environments, which further requires first an information acquisition system and then action recognition system. In our proposed solution, we divide the task into four major steps: (1) Feature sensing and enrichment via an IoT platform; (2) Automatic intelligent status labelling; (3) Status segmentation and feature extraction; and (4) Clustered personalized learning. We will introduce the details in the following subsections.

### *Feature sensing and enrichment via IoT platform*

By using Internet of Things (IoT) and cloud- and cognitive-based analytics technologies, we have built an intelligent vehicle surrounding information sensing system, as shown in the first three modules in the framework in Figure 1: Vehicle Sensor Readings from SQUAIR Mobile, Weather Data via ContextMap, and Air Pollution Data from Monitor Stations.

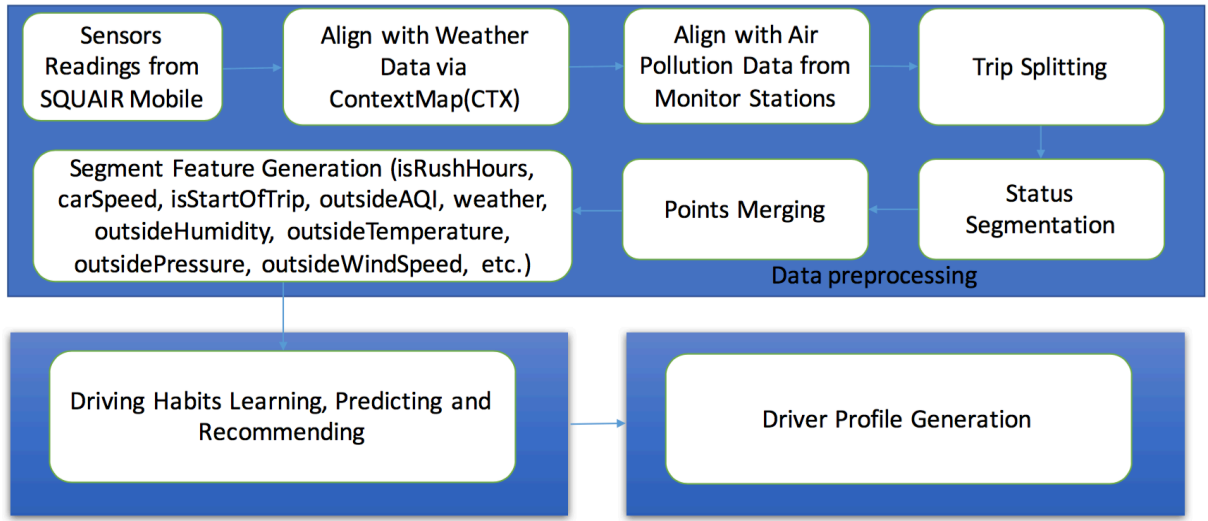

Figure 1. Overview of the system framework.

The data collected by the sensors are classified into three categories: environment data, vehicle running data, and traffic data. In the environment category, we collect weather, outside temperature and humidity, air quality index, outside pressure and wind speed etc.(need a complete list here) In the vehicle running category, we collect vehicle's running speed, GPS locations , in-vehicle PM2.5 (???), TVOC(Volatile Organic Cmpounds) etc. In the traffic feature category, we collect traffic time, whether it is the start of a trip, traffic congestion index (free, congestion, heavy congestion), etc. Some of the features, such as in-vehicle temperature/humidity or PM2.5 can be obtained directly from the sensors we employ, while the others, such as outside weather, PM2.5, wind speed, etc., need to be enriched from public dataset, for example, public air quality monitoring system or public weather prediction website.

*Automatic intelligent status labelling*

There are plenty of driver actions we are interested, however, at this stage we focus our aims to be three actions, i.e., closing the window, opening the window and turning on the air condition. These three actions are classical ones that will have significant influence on the in-vehicle climate---air quality if surrounding environment is polluted, in-vehicle temperature and humidity, etc. (The rationale of choosing these three actions can be updated if needed)

These three vehicle states, however, is not measurable directly. An intuitive approach is to manually record each driver's actions with time stamp. This is laborious and requires a lot of labeling work. Observing that there is a high correlation between the status and the in-vehicle feature changing, we design and implement an automatic status labeling algorithm.

The principles of our intelligent status labeling algorithms are based on the following observations: (1) when vehicle window closes, in-vehicle PM2.5 value will decrease, and TVOC value will increase; (2) when vehicle window opens, in-vehicle PM2.5 value will increase, TVOC value will decrease; (3) when the air condition is on, in-vehicle humidity will keep the same or decrease.

(One or more paragraph/diagram about this agorithms?)

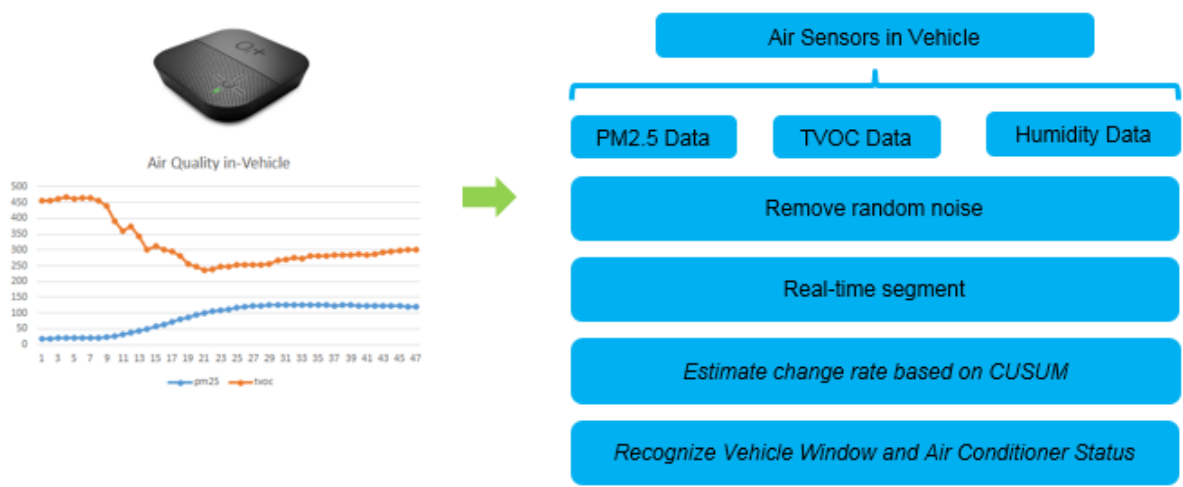

*Status segmentation and feature expansion*

Our feature is acquired at the frequency of 1Hz, and the automatic labeling algorithm can label each point's vehicle status. However, the purpose of our research is to identify each driver's action preference at a given environment in terms of a relative long and stable trip segment. So we need to merge the points with the same status and treat them as a training data unit, based on which we can further explore new features for training.

All the features mentioned in the first step are also sampled at the same frequency in real time, but for the purpose of recognizing the drivers' action pattern and the surrounding environment, in the learning stage, we shall group the points with the status together, and extract new features based on the grouping results.

Many of the original features can be still used in the feature expansion stage. These features are usually stable timewise or distribute relatively evenly in a trip-wise distance, for example, outside whether, temperature, humidity, wind speed etc. Many other features, though not directly observable from the sensors, can be extracted with our feature expansion approaches. These features include isRushHours (whether it is within traffic rush hours), carSpeed (derived from GPS location shift), isStartOfTrip (derived from trip time stamp), etc.

*Clustered personalized learning*

With preprocessed driver data, the actions status of which are automatically labeled and the features of which are extracted, we can analyze the relationship between driver's action and surrounding environmental features. Different drivers may have distinguished action patterns, thus accumulating all the drivers' data together and train a general predictor usually will not work. This is also illustrated in the real data experiment section.

For each driver, we first separate all the driving data into different trips. For example, driving from home to work place may consider to be a valid trip. Within each trip, we extract segments with three different action status and train predicting model. According to our observation of this application, different features play different importance significance while deciding a segment's action status. To utilize this nature, we explored different classification algorithms ---SVM, Decision tree, Knn, Ensemble boosted trees, and Gradient Boosting Decision tree--- and find that Gradient Boosting Decision Tree achieves the best predicting results.

Gradient boosting is a machine learning technique for classification problem which produce a prediction model in the form of an ensemble of week prediction models, typical decision trees, by building models in a stage-wise fashion and then generalizing them via optimizing an arbitrary differentiable loss function [4]. It can be formalized as follows.

$$F_0(x) = \arg\min_{\gamma} \sum_{i=1}^{n} L(y_i, \gamma),$$

$$F_m(x) = F_{m-1}(x) + \arg\min_{f \in \mathcal{H}} \sum_{i=1}^{n} L(y_i, F_{m-1}(x_i) + f(x_i))$$

In the future, as the driver number scale to hundreds of thousands of people, for the computational efficiency, we can also instead of training the driver model one-by-one, we can first cluster the drivers, and then training a model for each and every cluster.

Also, after each driver's different action states are classified, we can extract the drivers' preferences. For example, we can evaluate the most likely preferable in-vehicle temperature by finding the most frequently used segment steady values while the air condition is on. These values (temperature, humidity, etc.) can be widely used in the advanced assistive driving systems or next generation self-driving cars.

**Real data experiments**

We have collected data from 9 drivers driving in multiple cities worldwide for in total 320 hours and around 11,370 kilometres. We will illustrate our results from three perspectives. First, we will introduce the data we collect. Then, we will analyse the correlation between the surroundings and in-vehicle air quality under different driver actions, based on which we train the prediction models for new incoming surrounding information. Finally, we will talk about the personalized driver preference calculation and profile generation.

*Data Overview*

We obtained the in-vehicle temperature, humidity, PM2.5 features using SQUAIR Mobile air purifier, as shown in Figure 2(a), and collect the data via our iOS App AIR(https://appsto.re/us/mlvZdb.i), the interface of which is shown in Figure 2(b).

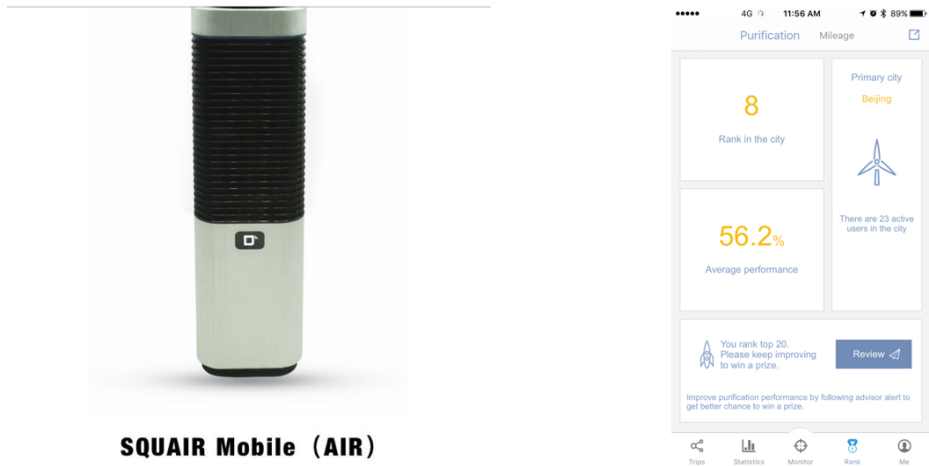

Figure 2. (a)Mobile air purifier          (b) Data collection App interface

Figure 3(a) shows each driver's time spent in each and every one of the three action states and Figure 3(b) shows the ratio of the three states.

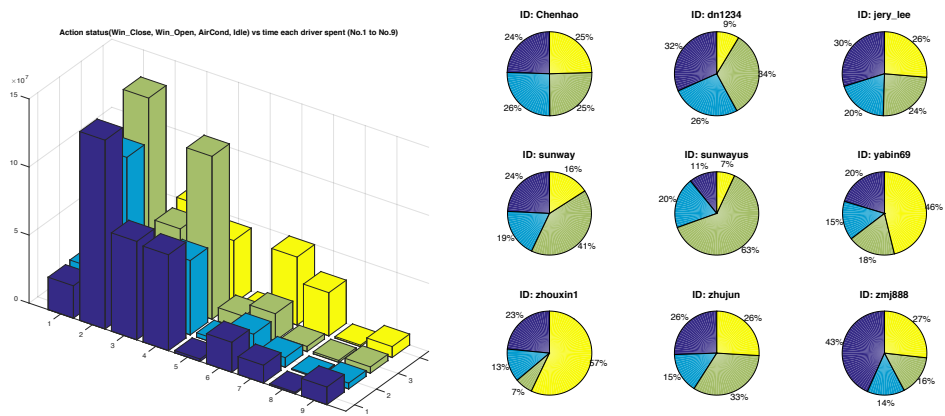

Figure 3. (a) Time spent by each driver. (b) Three actions' ratio for each driver

Different color represents different actions. Purple for closing window, blue for opening window and yellow for turning on the air condition, and yellows stands for the states that we are not sure about. The height of the bars in Figure 3(a) stands for the amount of time the corresponding driver spends in the current states within the database. In the Figure 3(b), different driver has different ration in spending time among these three actions.

*Correlation analysis and prediction model*

For better understanding the relationship under different actions between the environment and in-vehicle air quality, which could be one of the important climate we want to control especially driving in heavily air polluted cities, we analyzed the air quality level against in-vehicle temperature, humidity, car speed, pressure and window speed.

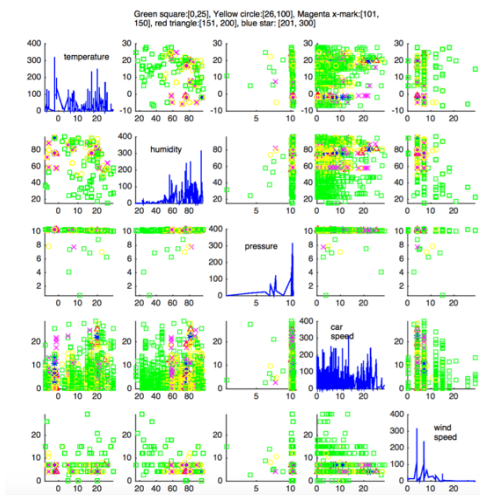

(a) Air quality correlation matrix in closing window state

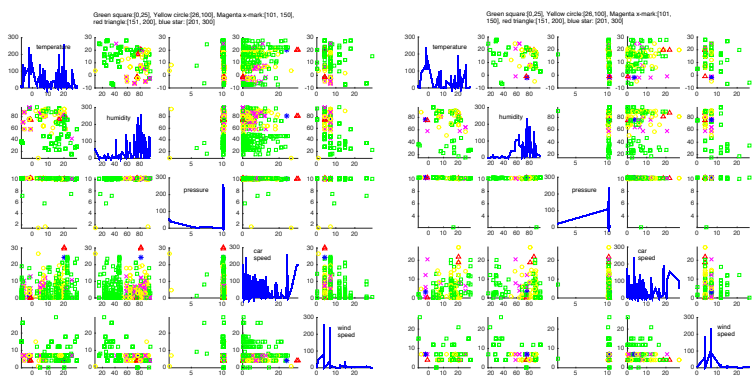

(b) Correlation matrix in opening window (c) Correlation matrix in air condition on
Figure 4. Correlation matrix between in-vehicle air quality and surroundings under different actions.

Figure 4 shows our results of the correlation matrix between impact factors and the in car PM2.5 under different action status---closing window, opening window and air condition on. There are many interesting observations. For example, from the figure of the second column and first row of Figure 4(a), we can see that high pollution usually happens in the condition when the outside humidity is very high, especially when the temperature is also high, based on which we can alert the drivers to pay special attention for driving under such conditions. As another example, the green square points ratio, which represents the trip segments that have in-vehicle PM2.5 less than 25, when the air condition is on (right figure) is much higher than the condition where the window is open (left figure). This indicates that turning on the air condition can reduce the PM2.5 exposures, and thus is advised when outside air quality is poor, while opening the window in such environment is highly not recommended.

To utilize this observation, we therefore trained a couple of classification predictors for the four major drivers' data. Figure 5 illustrates our prediction accuracy under different models. Gradient boosting decision tree achieves the highest accuracy among the models, since it is more fit to the nature of influencing process between the surrounding factors and the final in-vehicle air quality.

| Algorithm/ID | dn1234 | Jery_lee | sunway | chenhao |
|---|---|---|---|---|
| Linear SVM | 84.7% | 82.7% | 70.2% | 67.3% |
| Decision Tree | 86.1% | 84.5% | 79.5% | 67.3% |
| Knn | 85.4% | 82.7% | 68.9% | 59.6% |
| Ensemble Boosted Trees | 85.0% | 83.6% | 62.3% | 67.3% |
| **GBDT** | **98.95%** | **86.49%** | **84.0%** | **72.22%** |

Figure 5. Driver action prediction training models.

The above result shows that there is a high correlation between the surrounding factors and a driver's action, which is consistent with our previous analysis in Figure 4.

*Personalized preference calculation*

Different drivers have distinguishable preferences while driving depending on a couple of factors, such as economic views, sensitive to coziness, personal habits, etc. Figure 6 shows our experiment in a few of our drivers.

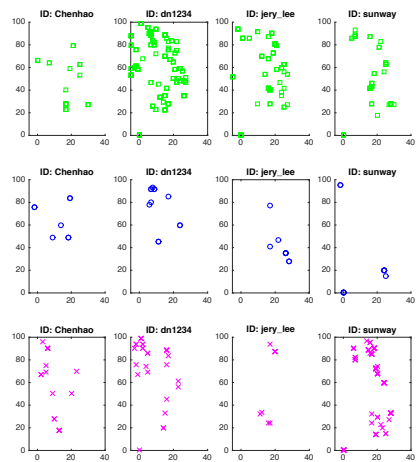

Figure 6. Driver pattern recognition.(?)

There are some interesting findings. Some drivers have unique preference in choosing actions under different environment. For example, the fourth driver (ID: sunway) spends very few time in opening the window comparing with others (Row 2, Column 2 (R2, C2)). Also, the driver usually turn on the air condition when temperature is great than 20 degree (R3, C4), and will turn off the air condition but close the window when temperature is between 10 to 20 degree (R 3, C4). We can therefore use the data to recognize the drivers' preference, based on which we can therefore providing IC-AQC action recommendation. When humidity is high, drivers tends to either turning on the air condition (R3, C2) or open the window (R2, C2). Some drivers tends to open the window when the humidity is high (e.g. ID: dn1234 in R2,C2 and ID: Chenhao in R2, C1), while some others tend to open the window while humidity is comfortable (e.g. ID: jery_lee in R2 C3).

When the driver number scales up, it is not practical to generate profile by checking the data one-by-one. We need an automatic profile generating approach. As an example, in Figure 7, we divide the temperature and humidity into three levels respectively---high, middle and low, and use the amount of time each driver spent in these categories in each of the three actions as the descriptor of a driver. Figure 7(a) gives an overview for all the drivers, and Figure 7(b) shows the descriptor for four drivers.

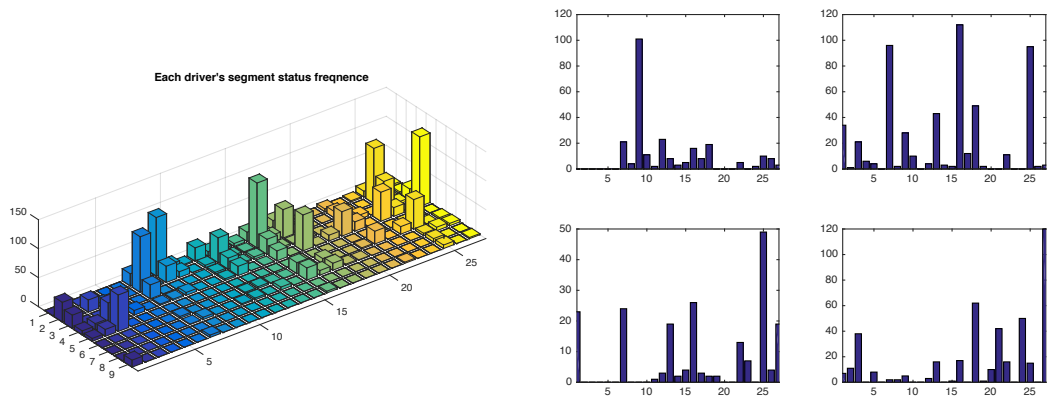

Figure 7. (a) Descriptors overview    (b) Descriptor vector for four drivers

After we group the drivers using the above generated profile, we can also further determine quantitative features of each driver. For example, Figure 8 shows the temperature selection of two drivers while air condition is on.

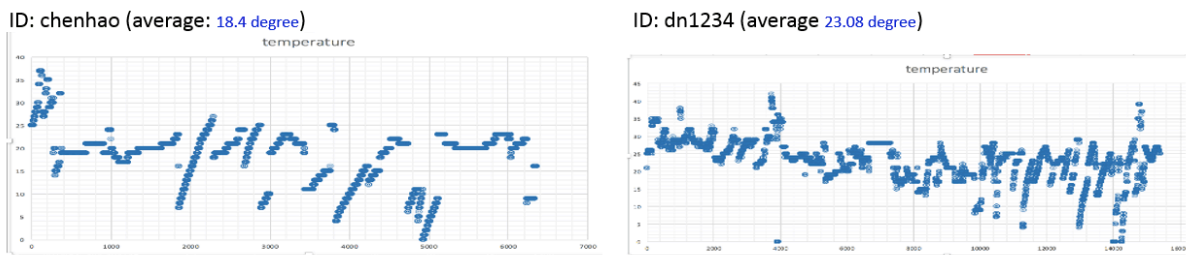

Figure 8. Drivers' preference difference in in-vehicle temperature while air condition is on

From the above figure, we can provide action preference profile for each and every driver. These profiles can therefore be used for providing personalized service in advanced driving assistance system or future next generation self-driving cars.

## Conclusion and Future Work

In this paper, we introduce a new driver climate control behaviour recognition and recommendation system for advanced driving assistance in the next generation self-driving vehicles. We describe the three key components of the system: vehicle and environment data collection, driver behavior and vehicle status detection, and personalized driver habits learning and preference recommendation. Real driving data of 320 hours (11,370km) from multiple cities worldwide are utilized to demonstrate the effectiveness of our solution. In our ongoing work, we are extending the driver numbers from current few drivers to around 500 drivers, and we are also expecting more data to come in to generate more interesting results. Also, we would like to extend the analysis from using just current data to using both new data and data of previous states for improving the prediction accuracy as well as performing more interesting driver pattern recognition.

In the future, the module of knowledge of personalized climate preference can be used as an API and integrated with the control package of an unmanned vehicle, aiming to improve the friendliness and comfort of self-driving.